\documentclass[runningheads]{llncs}

\usepackage{graphicx}
\usepackage{bm}
\usepackage{xcolor}

\usepackage{booktabs}

\begin{document}
\title{An End-to-End Approach for Seam Carving Detection using Deep Neural Networks\thanks{The authors are grateful to S\~ao Paulo Research Foundation (Fapesp) grants \#2013/07375-0, \#2014/12236-1, \#2019/07665-4, to the Petrobras grant \#2017/00285-6, to the Brazilian National Council for Research and Development (CNPq) \#307066/2017-7 and \#427968/2018-6, as well as the Engineering and Physical Sciences Research Council (EPSRC) grant EP/T021063/1 and its principal investigator Ahsan Adeel. }}

\titlerunning{Seam Carving Detection using Deep Neural Networks}

\author{Thierry P. Moreira\inst{1}\orcidID{0000-0002-3410-6247} \and Marcos Cleison S. Santana\inst{1}\orcidID{0000-0003-2568-8019} \and Leandro A. Passos\inst{2}\orcidID{0000-0003-3529-3109} \and Jo\~ao Paulo Papa\inst{1}\orcidID{0000-0002-6494-7514} \and Kelton Augusto P. da Costa\inst{1}\orcidID{0000-0001-5458-3908}}

\authorrunning{Moreira et al.}

\institute{Department of Computing, S\~ao Paulo State University, Av. Eng. Luiz Edmundo Carrijo Coube, 14-01, Bauru, 17033-360, Brazil
\email{\{thierrypin,marcoscleison.unit\}@gmail.com}, \{joao.papa, kelton.costa\}@unesp.br \and CMI Lab, School of Engineering and Informatics, University of Wolverhampton, Wolverhampton, United Kingdom
\email{l.passosjunior@wlv.ac.uk} }
\maketitle              
\begin{abstract}
Seam carving is a computational method capable of resizing images for both reduction and expansion based on its content, instead of the image geometry. Although the technique is mostly employed to deal with redundant information, i.e., regions composed of pixels with similar intensity, it can also be used for tampering images by inserting or removing relevant objects. Therefore, detecting such a process is of extreme importance regarding the image security domain. However, recognizing seam-carved images does not represent a straightforward task even for human eyes, and robust computation tools capable of identifying such alterations are very desirable. In this paper, we propose an end-to-end approach to cope with the problem of automatic seam carving detection that can obtain state-of-the-art results. Experiments conducted over public and private datasets with several tampering configurations evidence the suitability of the proposed model.
\end{abstract}

\section{Introduction}
\label{s.introduction}

Seam Carving is an image operator created for content-aware image resizing, and it works by searching for the least relevant pixel paths (seams) in the image so that, when they are removed or duplicated, the figure size is changed without distortion or relevant information loss. Roughly speaking, seams are defined as a connected line of pixels that crosses the entire image vertically or horizontally.

The general effect for the human eye is that the image was already captured that way, i.e., without being ``distorted".  Seam carving has good uses for personal editing, but fake images may be used for illegal purposes too, such as perjury, counterfeiting, or other types of digital forgery. This scenario fostered the need to ensure the image integrity, primarily, but not exclusively, for forensic analysis. Therefore, developing efficient methods for its detection became essential.

Avidan and Shamir~\cite{avidan2007} originally employed the concept of object removal using seam carving. In a nutshell, the idea consists in driving all seams to pass through the object of interest. One may reapply seam carving to enlarge the image to its original size. Seam carving may highlight some texture artifacts in the resulting image as well~\cite{nataraj2021seam,li2020identification,nam2020deep}, which can be considered for further automatic detection. Although these artifacts are mostly imperceptible to the human eye, they can be detected by computer vision techniques. This effect was explored by Yin et al.~\cite{yin2015}, which applied a set of descriptors directly on the LBP (Local Binary Descriptors) image~\cite{wang1990}. The authors employed eighteen energy information and noise-based features from~\cite{ryu2014}, and defined six more features based on statistical measures from the seams.

Zhang et al.~\cite{zhang2017} proposed an approach that consists of extracting histogram features from two texture descriptors, i.e., LBP and the Weber Local Descriptor~\cite{chen2010}. The methodology first extracts both descriptors from the image, and then their histograms are merged for the further application of the Kruskal-Wallis test for feature selection purposes~\cite{saeys2007}. The resulting vector is then classified with the well-known Support Vector Machines (SVM) technique~\cite{CortesML:95}. Their results indicated that texture features have a strong descriptive power for the problem at hand.

Avidan and Shamir~\cite{avidan2007} originally employed the concept of object removal using seam carving. In a nutshell, the idea consists in driving all seams to pass through the object of interest. One may reapply seam carving to enlarge the image to its original size. Seam carving may highlight some texture artifacts in the resulting image as well~\cite{nataraj2021seam,li2020identification,nam2020deep}, which can be considered for further automatic detection. Although these artifacts are mostly imperceptible to the human eye, they can be detected by computer vision techniques. This effect was explored by Yin et al.~\cite{yin2015}, which applied a set of descriptors directly on the LBP (Local Binary Descriptors) image~\cite{wang1990}. The authors employed eighteen energy information and noise-based features from~\cite{ryu2014}, and defined six more features based on statistical measures from the seams.


While seam carving detection has been approached using texture features, in this work we propose an end-to-end approach based on deep neural networks. We observe that feeding neural networks with the original images allows for better accuracies when the severity of the tampering is quite subtle to be detected. We show results that outperform the recent ones presented by Cieslak et al.~\cite{Cieslak18} on the smallest rates of change. Therefore, the main contribution of this work is to propose an end-to-end approach for seam carving detection on unprocessed images. This approach is more effective on finding small and subtle tempering.

The remainder of this paper is organized as follows. Section~\ref{s.theoretical} presents a theoretical background related to seam carving, Section~\ref{s.proposed} discusses the proposed approach. Sections~\ref{s.methodology} and~\ref{s.experiments} describe the methodology and the experimental section, respectively. Conclusions and future works are stated in Section~\ref{s.conclusions}.

\section{Theoretical Background}
\label{s.theoretical}
\newcommand\abs[1]{\left|#1\right|} 

In this section, we present the theoretical background concerning the main concepts of seam carving, as well as some energy functions commonly used to compute vertical/horizontal seams.

Seam carving is a content-aware image resizing approach that consists mainly of inserting or removing seams iteratively until the image achieves a desired width or height. Roughly speaking, a seam is defined as a connected line of pixels in completely crossing the image. Seams are constructed using an energy function to ensure that only the low-energy pixels will be removed, keeping the primary structure of the image. These low-energy pixels belong to a low-frequency region in the image where the changes might be imperceptible.

One of the crucial aspects concerning seam carving stands for the concept of ``energy", which is used to define the ``paths to be carved"\ in the image. A standard energy function \emph{e} applied to an image $\bm{I}$ can be defined as follows: 
\begin{equation}
\label{eq.energy}
e(\bm{I}) = \abs{\frac{\partial}{\partial x} \bm{I}} + \abs{\frac{\partial}{\partial y} \bm{I}},
\end{equation}

Such a formulation can also be extended to pixels, i.e., one can compute the energy at pixel $I_{x,y}$. Roughly speaking, a vertical seam is an $8$-connected path of adjacent pixels from the top to the bottom of the image, containing only one pixel from each row of the image. An analogous definition can be derived for horizontal seams.

Moreover, the energy function applied to a seam $\bm{s}$ is defined as follows:
\begin{equation}
\label{e.seam2}
E(\bm{s}) = \sum_{i=1}^{m}{e(s_{i})},
\end{equation}
where $e(s_{i})$ denotes the partial energy computed at the $i$-th pixel of seam $s$. The optimal seam, which is the one that minimizes Equation~\ref{e.seam2}, can be efficiently computed using dynamic programming.

\begin{figure*}[!htb]
      \centering
     \begin{tabular}{cc @{\hskip .7cm} cc}
     
	\multicolumn{2}{c}{Resizing} & \multicolumn{2}{c}{Object Removal} \\
	
	\multicolumn{2}{c}{\includegraphics[width=3cm]{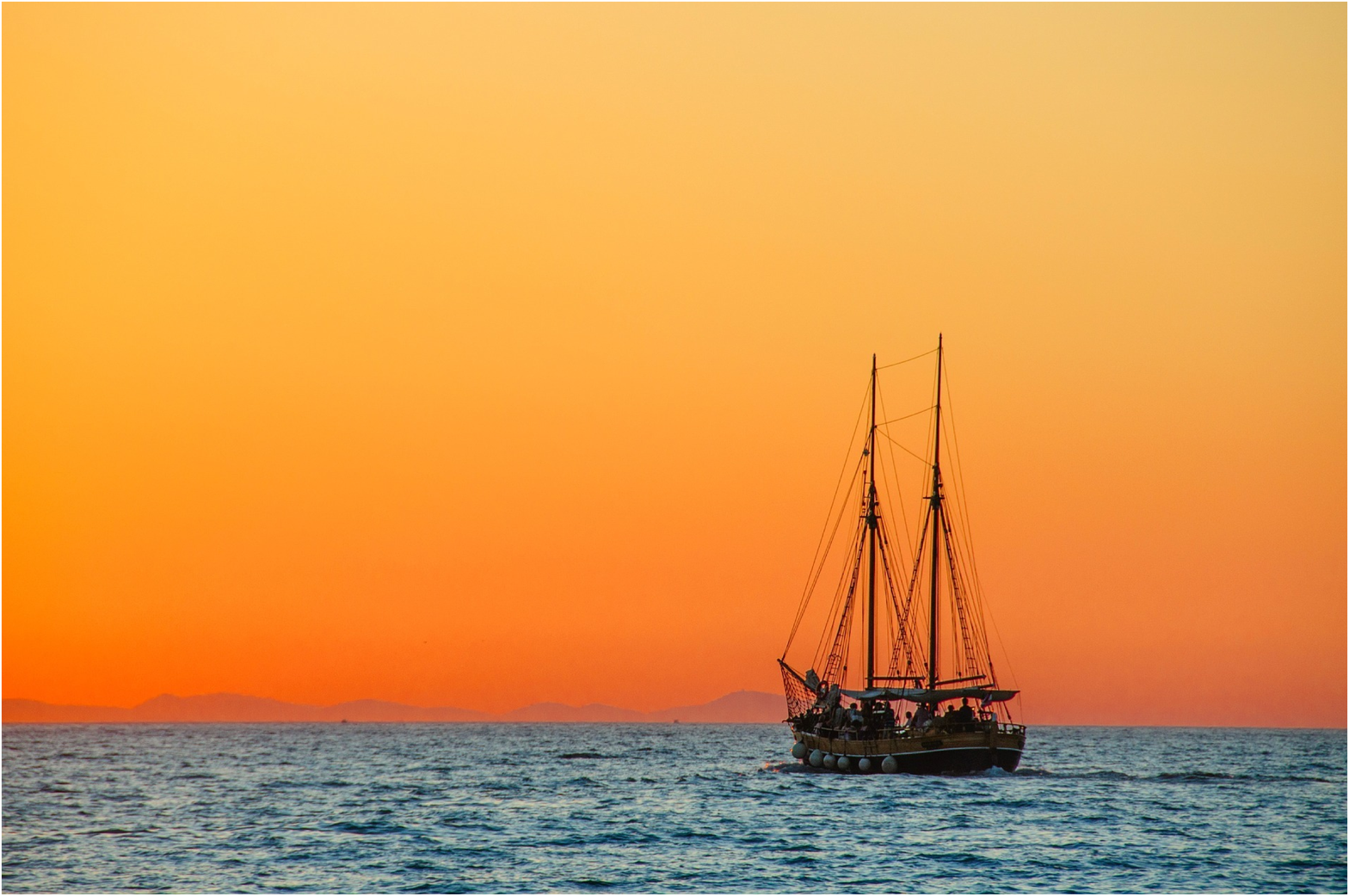}} & \includegraphics[width=3cm]{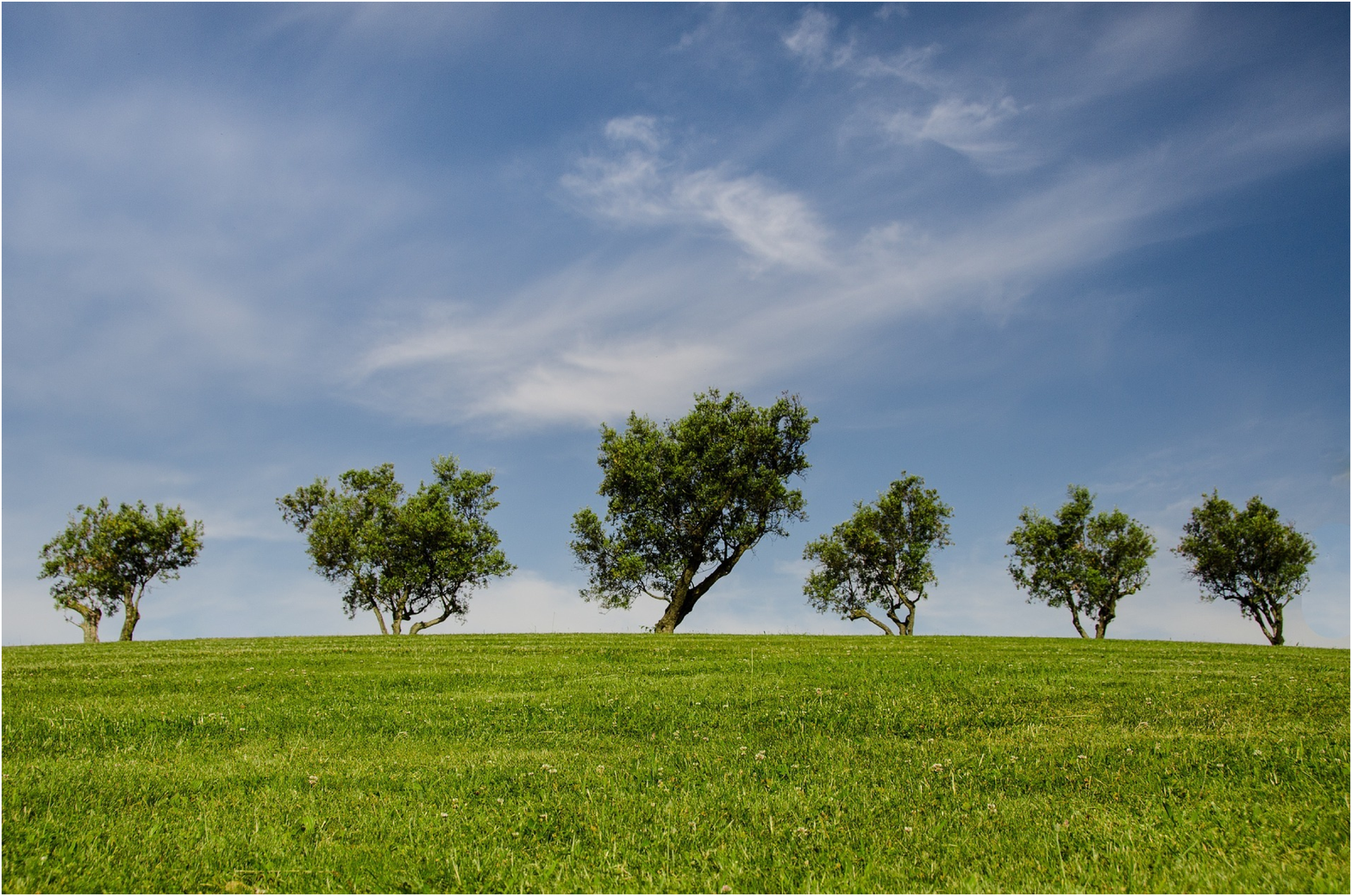} & \includegraphics[width=3cm]{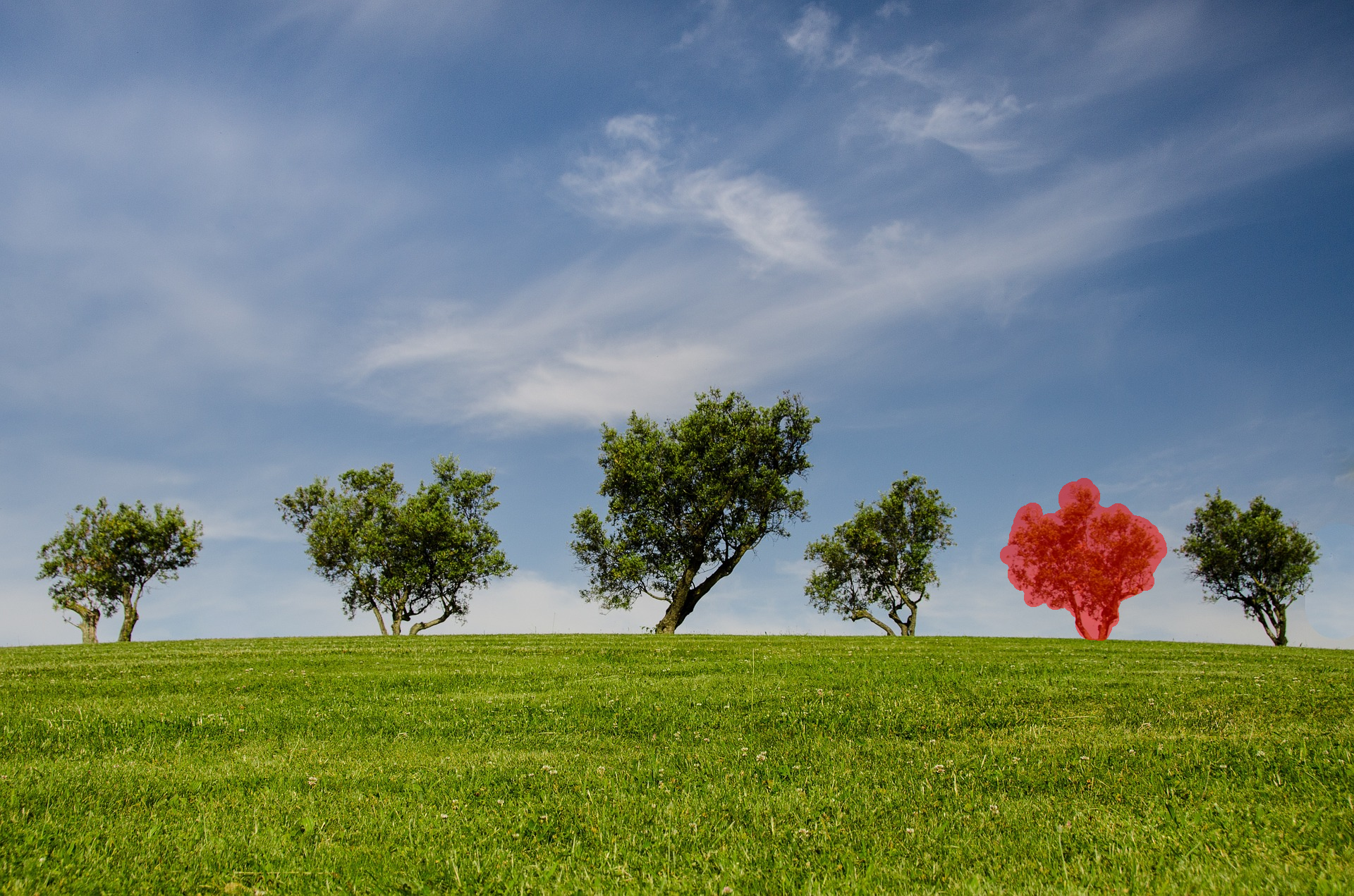} \\
	\multicolumn{2}{c}{(a)} & (d) & (e) \\
     
 	\includegraphics[width=2.75cm]{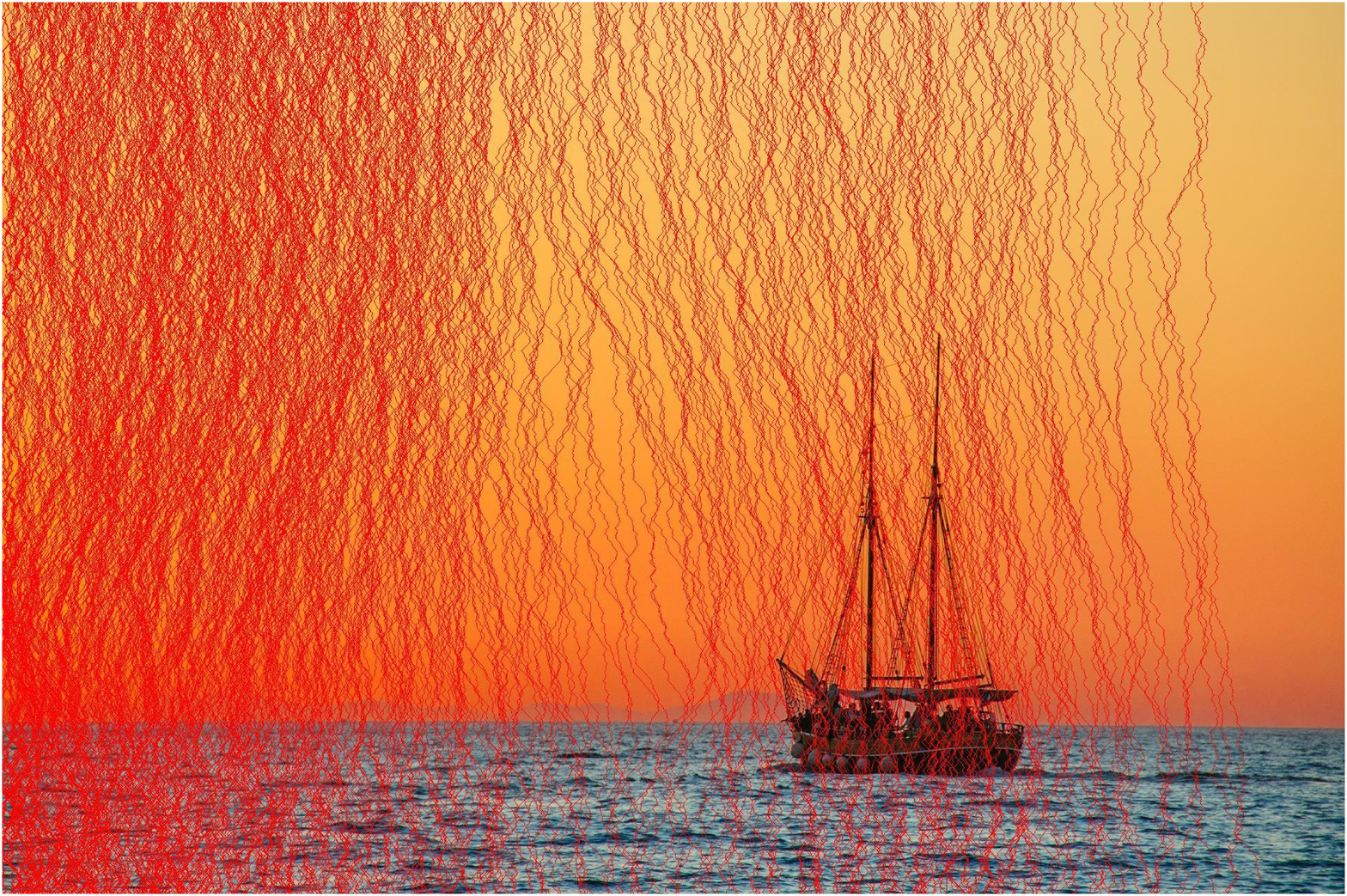} & \includegraphics[width=1.83cm]{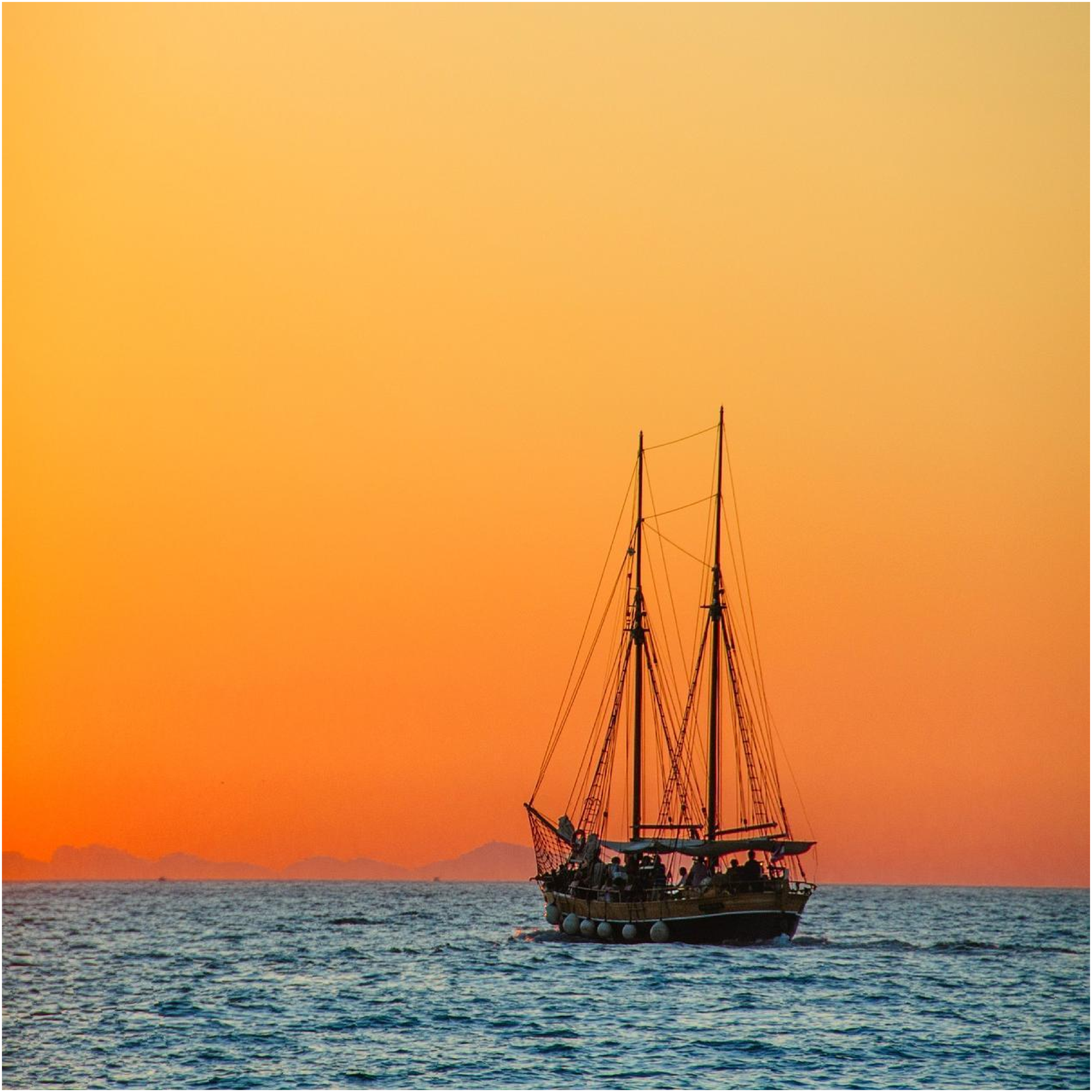} & \includegraphics[width=3cm]{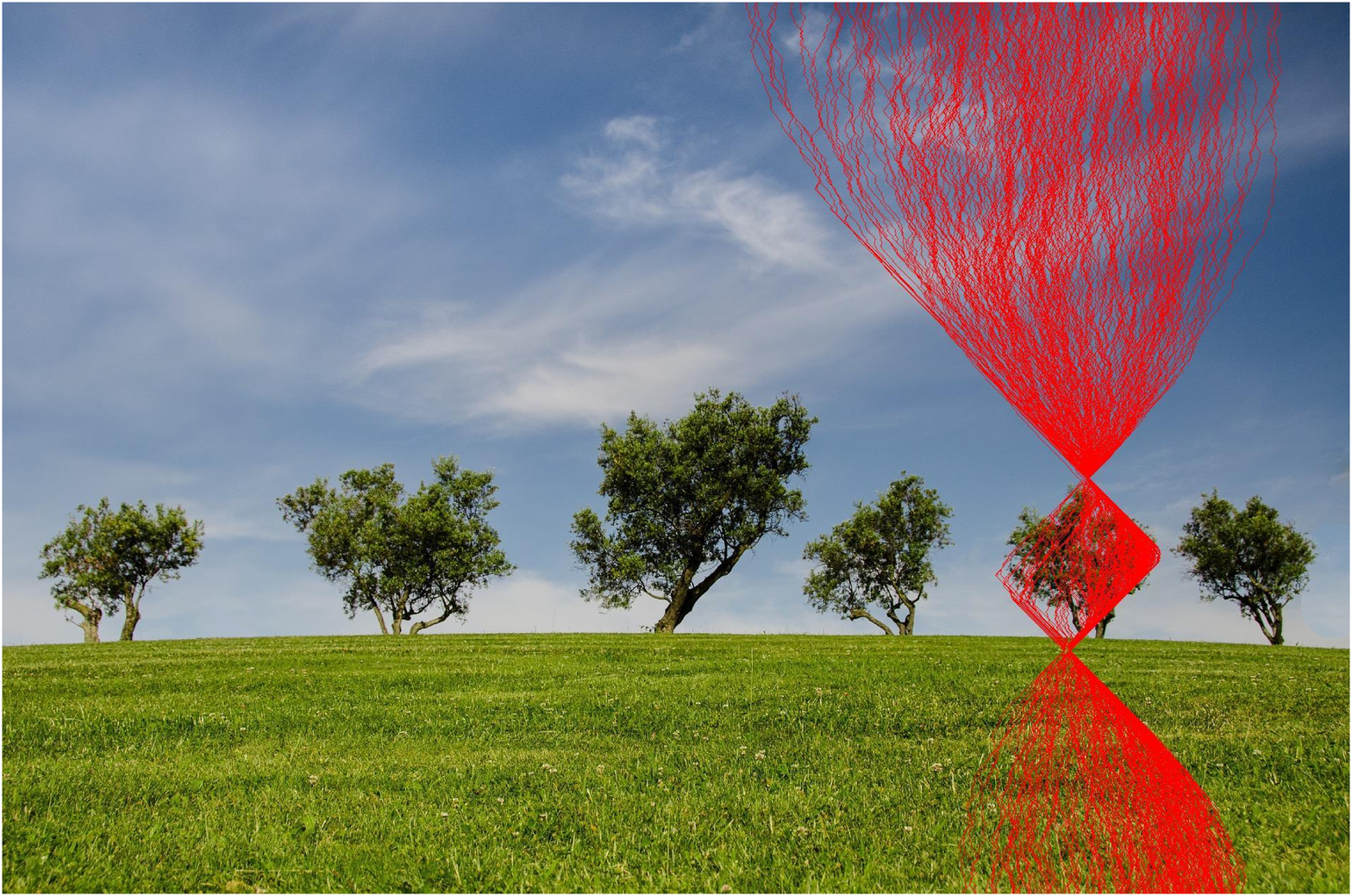} & \includegraphics[width=2.64cm]{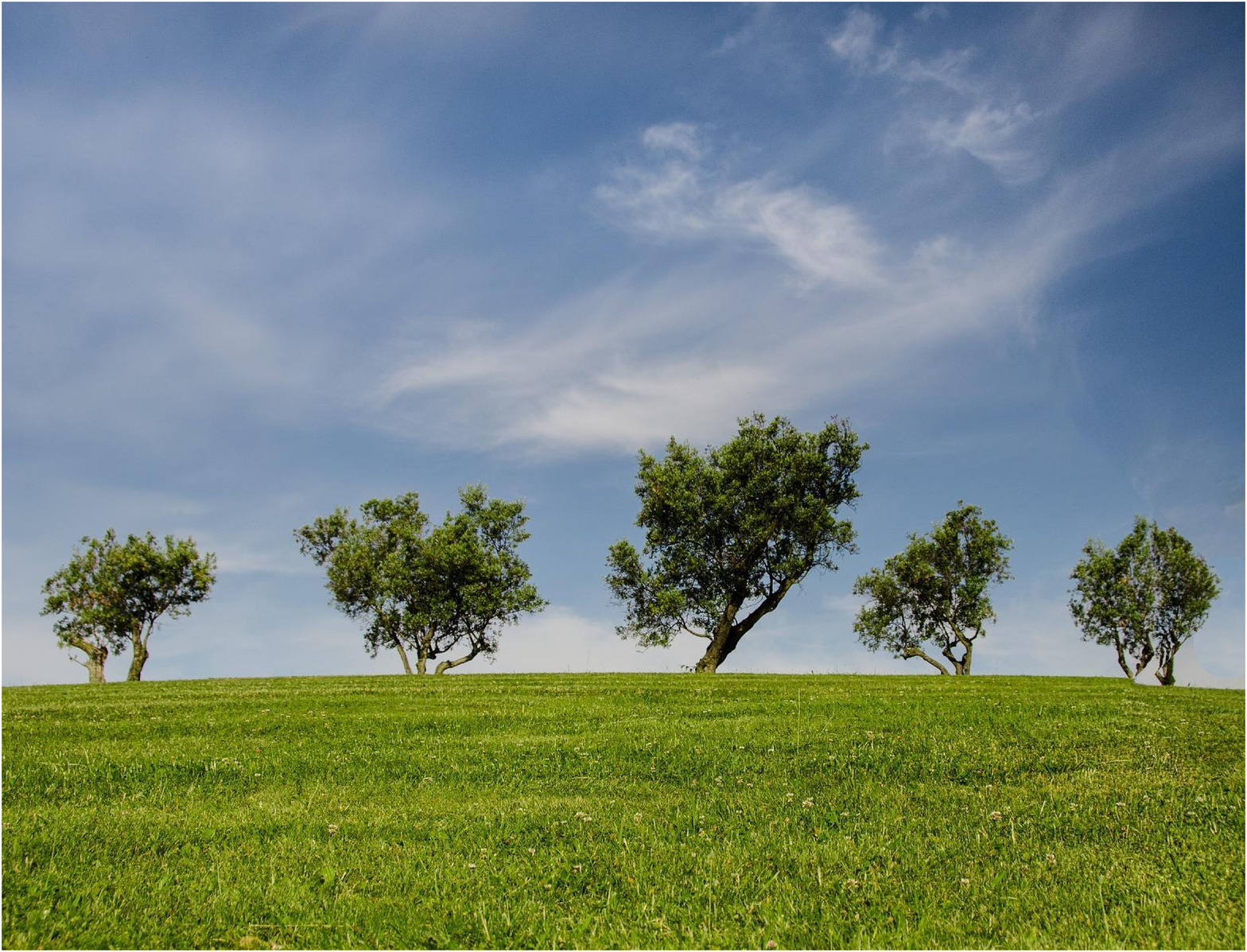} \\
 	(b) & (c) & (f) & (g) \\
     \end{tabular}	
     \caption[Caption for LOF petro]{Left: Seam carving for resizing: (a) original image, (b) original image with seams marked for removing, and (c) resized image using seam carving. Right: Seam carving for object removal: (a) original image, (b) original image with object marked for removal, (c) original image with seams marked for object removal, and (d) desired object removed using seam carving.}
   \label{f.sc_examples}
\end{figure*}

Figure~\ref{f.sc_examples} (left) demonstrates the process of seam carving over Figure~\ref{f.sc_examples}(a), where Figure~\ref{f.sc_examples}(b) shows the seams computed in the vertical axes, and Figure~\ref{f.sc_examples}(c) shows Figure~\ref{f.sc_examples}(a) after removing those seams. We can also use seams to remove objects from an image, thus restraining the search of seams in a given space. Figure~\ref{f.sc_examples} (right) depicts an example of such process\footnote{Figure~\ref{f.sc_examples} is licensed under Creative Commons 0 Public Domain.}.

Avidan and Shamir~\cite{avidan2007} propose other energy functions for the problem (for example, Histogram of Oriented Gradients (HoG)) and argue that no single formulation works well for all images at once, but they all may have similar results in the end. The different formulations may vary in the rate at which visual artifacts are introduced.

\section{Proposed Approach}
\label{s.proposed}

As mentioned before, removing or inserting a seam leaves intrinsic local patterns. This way, the inspection of the seam carving traces can be accomplished by exploring the local discontinuities of pixel values. While these discontinuities can be detected by texture-based feature extractors, a deep convolutional model is also capable of identifying such artifacts and distortions. To cope with such issues, we propose an end-to-end deep network training methodology to inspect the use of seam carving manipulation in images. 

We opted to employ the well-known deep neural network Xception since it innovates on using separable convolutions as the fundamental operator. The network is composed of three essential parts: (i) the entry flow, (ii) some repeating blocks forming the middle flow, and (iii) the exit flow. The output of exit flow can be used to feed dense layers and later on a classifier, while the entry flow is in charge of extracting low-level features with its residuals obtained from $1\times 1$ convolutions. The middle flow, with its repeating blocks of depthwise convolutions with simple residuals connections, extracts deeper features.

Other architectures could be employed -- the choice of the Xception architecture was motivated by the fact the network presents a good trade-off between representation ability and execution time. Such skill is due to its depthwise separable convolutions, which speeds up its execution, even when the number of parameters is kept the same. The advantages of the Xception can be observed in some previous works, such as Giancardo et al.~\cite{giancardo2018}, which employed the pre-trained model for brain shunt valve recognition on magnetic resonance images, and Liu et al.~\cite{liu2018}, that applied Xception for facial expression feature extraction.

\begin{figure*}[!htb]
   \centering
   \includegraphics[width=0.9\linewidth]{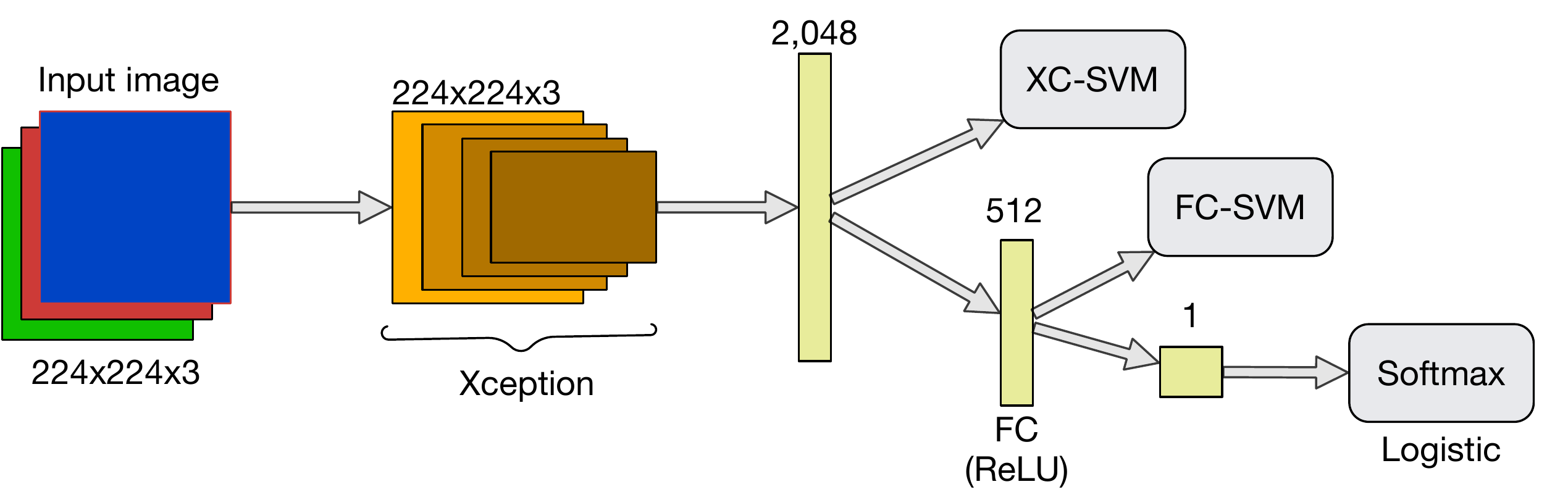}
   \caption{Proposed deep network architecture: two fully-connected layers are added right after the original Xception model to perform fine-tuning for seam carving recognition purposes.}
   \label{f.proposed_approach}
\end{figure*}

The model proposed in this work is based upon the original Xception architecture but with its original top layers (fully connected - FC) replaced by two new ones. The first FC layer reduces the dimensionality from 2,048 to 512, while the second is in charge of classifying the input image as seam carved or untouched. Figure~\ref{f.proposed_approach} depicts the architecture proposed in this work.
 
Training a CNN to a classification task opens the possibility of using intermediate CNN layers as a feature extractor to feed other classifiers. In our experiments, the aforementioned Xception architecture is stacked with fully-connected layers for fine-tuning purposes and weights learned over the well-know ImageNet dataset~\cite{SUTSKEVER}. When the training process converges, the last FC layer acts as a classifier, but one can remove a number of layers from the stack and use intermediate representations with other classification algorithms.

The model depicted in Figure~\ref{f.proposed_approach} allows us to observe three distinct descriptions from the same network: (i) the Xception output with size as of 2,048, (ii) the first fully-connected output with size as of $512$, and (iii) the classification output, which stands for a softmax layer activated with a Sigmoid function to scale the output between $0$ (not tampered image) and $1$ (seam-carved image). The loss function adopted in this work concerns the well-known ``binary cross-entropy". 

\section{Methodology}
\label{s.methodology}

In this section, we present the datasets and the experimental setup used to validate the robustness of the proposed approach for seam carving detection using deep networks.

\subsection{Datasets}
\label{ss.datasets}

In this paper, we used two datasets, being the former composed of natural images, from now on called ``SC-Image", and provided by Liu et al.~\cite{liu2013}. Such a dataset contains $5,150$ images cropped to $224\times224$ pixels and compressed with JPEG using a $75\%$ compression rate. Each image has $10$ modified counterparts, obtained by varying the rate of change by $3\%$, $6\%$, $9\%$, $12\%$, $15\%$, $18\%$, $21\%$, $30\%$, $40\%$, and $50\%$. The higher the rate of change, the easier for an algorithm to detect the seam carving tampering. Evaluation is performed using a 5-fold cross-validation scheme.

The other dataset is composed of images obtained from off-shore well drilling rigs during operation of Petrobras SA, a Brazilian oil and gas company. The dataset, hereinafter called ``Petrobras"~\cite{santanaIEEE-IS:19}, is private and therefore not available publicly. For each rate of change, we generated $564$ images, thus summing up a dataset composed of $6,204$ images. Similarly to the other dataset, evaluation is performed using a 5-fold cross-validation scheme.

\subsection{Experimental Setup}
\label{ss.setup}

To fine-tune the network, we used as training set the original images together with their ten tampered versions. To avoid biasing the network, we balance each input batch so that one half of is positive and the other half is negative. Therefore, the resulting network is less invariant to the proportion of changes.

We considered four approaches for data augmentation purposes, each with a $50\%$ chance of occurrence for a given image: (i) horizontal flip, (ii) vertical flip, (iii) Gaussian noise of zero mean and standard deviation as of $0.2$ (the original pixel values are scaled to $[-1,1]$), and (iv) a random black column with width of $10$ pixels.

The training process was performed using the Stochastic Gradient Descent with Restart (SGDR) optimization algorithm~\cite{loshchilov2016sgdr}. In this paper, the learning rate varied as a function of the cosine with an abrupt restart at the end of each cycle. The initial maximum learning rate was $lr_{max}=10^{-2}$, $lr_{max}$ value, decayed by $30\%$ at each cycle, with lower limit $lr_{min}=10^{-5}$. The initial period was composed of $10$ epochs. The training procedure was implemented with Keras 2.2.4~\cite{chollet2015keras} and Tensorflow 1.12.0 as a backend.

As depicted in Figure~\ref{f.proposed_approach}, we considered FC-SVM with 512 features, XC-SVM with 2,048 features, and the standard output of the CNN with a softmax layer, hereinafter called SM-CNN, for comparison purposes. Additionally, concerning dataset SC-Images, we also compared the proposed approaches against the works by Cieslak et al.~\cite{Cieslak18} and Zhang et al.~\cite{zhang2017}. Regarding the SVM implementation, we used the library provided by Fan et al.~\cite{liblinear} within SciKit-Learn~\cite{scikit-learn} package.

\section{Experiments}
\label{s.experiments}

In this section, we present the results obtained over SC-Image and Petrobras datasets, as well as a more in-depth discussion about the robustness of the proposed approaches.

\subsection{SC-Image Dataset}
\label{ss.cs_image_dataset}

Table~\ref{t.experiments_sc_image_dataset} presents the accuracy results concerning seam carving detection over the SC-Image dataset, where the best values are in bold. For lower levels of seam carving (i.e., rates of change), the proposed approaches obtained the best results with SM-CNN and XC-SVM models. Such low levels of distortion make it difficult for the network to discriminate between normal and tampered images since the differences are subtle, i.e., few and sparse information distinguish them. However, the proposed approaches were able to detect seam-carved images with suitable accuracy.

\begin{table*}[!htb]
  \centering
  \caption{Accuracies obtained over SC-Image dataset.}
  \label{t.experiments_sc_image_dataset}
  \begin{tabular}{rrrrrr}
  \toprule
                      & \multicolumn{5}{c}{Accuracy (\%)}          \\
  Rate of change      & SM-CNN         & FC-SVM  & XC-SVM   & Cieslak et al.~\cite{Cieslak18}  & Zhang et al.~\cite{zhang2017} \\ \midrule
  3\%                 & 84.03          & 85.05   & \textbf{85.19} & 81.46 & 68.26\\
  6\%                 & 87.96          & 87.76   & \textbf{88.02} & 83.50 & 84.99 \\
  9\%                 & \textbf{90.58} & 90.02   & 90.06          & 81.72 & 89.86 \\
  12\%                & \textbf{92.86} & 91.90   & 91.86          & 84.21 & 92.72 \\
  15\%                & 93.98 & 93.21   & 93.42          & 87.51 & \textbf{95.03} \\
  18\%                & 95.15 & 94.15   & 94.40          & 89.73 & \textbf{95.94} \\
  21\%                & 94.66          & 94.63   & 94.94 & 88.91 & \textbf{96.02}\\
  30\%                & 95.49          & 96.18   & 96.51 & 91.23 & \textbf{97.58}\\
  40\%                & 95.73          & 97.23   & 97.71 & 95.12 & \textbf{98.64}\\
  50\%                & 95.73          & 97.65   & 98.10 & 98.90 & \textbf{99.20}\\ \bottomrule
  \end{tabular}
\end{table*}
 
For intermediate values of tampering (i.e., $9\%$--$18\%$), the SM-CNN approach obtained the best results with a small margin regarding XC-SVM. Higher levels of tampering are easier to detect since the seam removal artifacts are present in abundance. For $50\%$ of reduction level, SM-CNN obtained $95.73\%$ of classification rate, while XC-CNN yielded $98.10\%$ of accuracy. Although our approch obtained smaller accuracies on higher rates of change compared with Zhang et al.~\cite{zhang2017}, it achieved much higher results on more subtle tempering.

To explore the decision mechanism that led to the results, we employed dimensionality reduction on the output features. The reduced space was computed using the Uniform Manifold Approximation and Projection (UMAP) technique~\cite{mcinnes2018} using $20$ neighbors for further projecting the output onto a 2D space. This projection method is based on  Riemannian geometry and algebraic topology. Figure~\ref{f.umap} (left) illustrates projections for SC-Image dataset images concerning different classifiers and rates of change.

\newcommand\height{2.5cm}

\begin{figure*}[!htb]
  \begin{tabular}{cc@{\hskip .3cm}cc}
\multicolumn{2}{c}{SC-Image} & \multicolumn{2}{c}{Petrobras} \\
\includegraphics[width=.24\linewidth,height=\height]{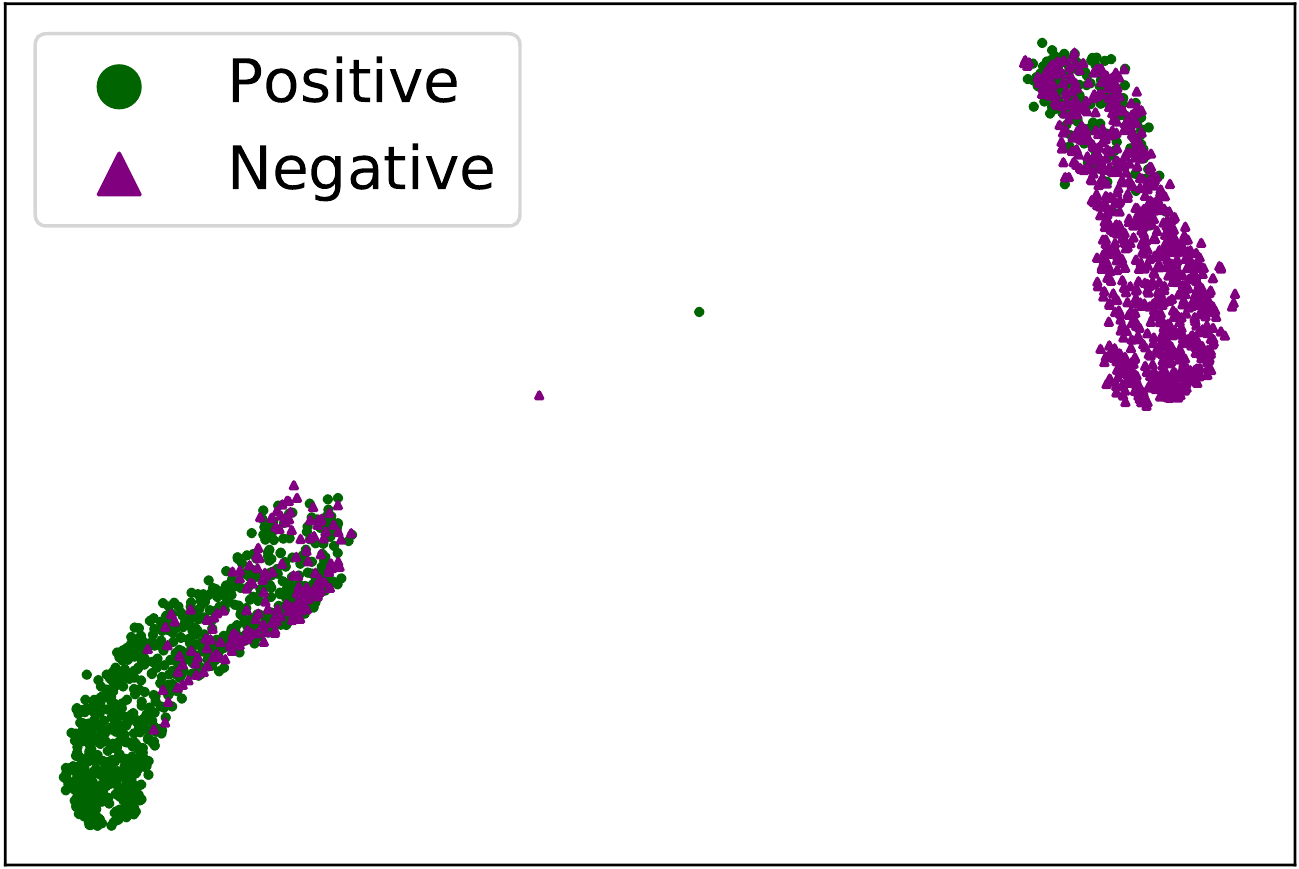} & \includegraphics[width=.24\linewidth,height=\height]{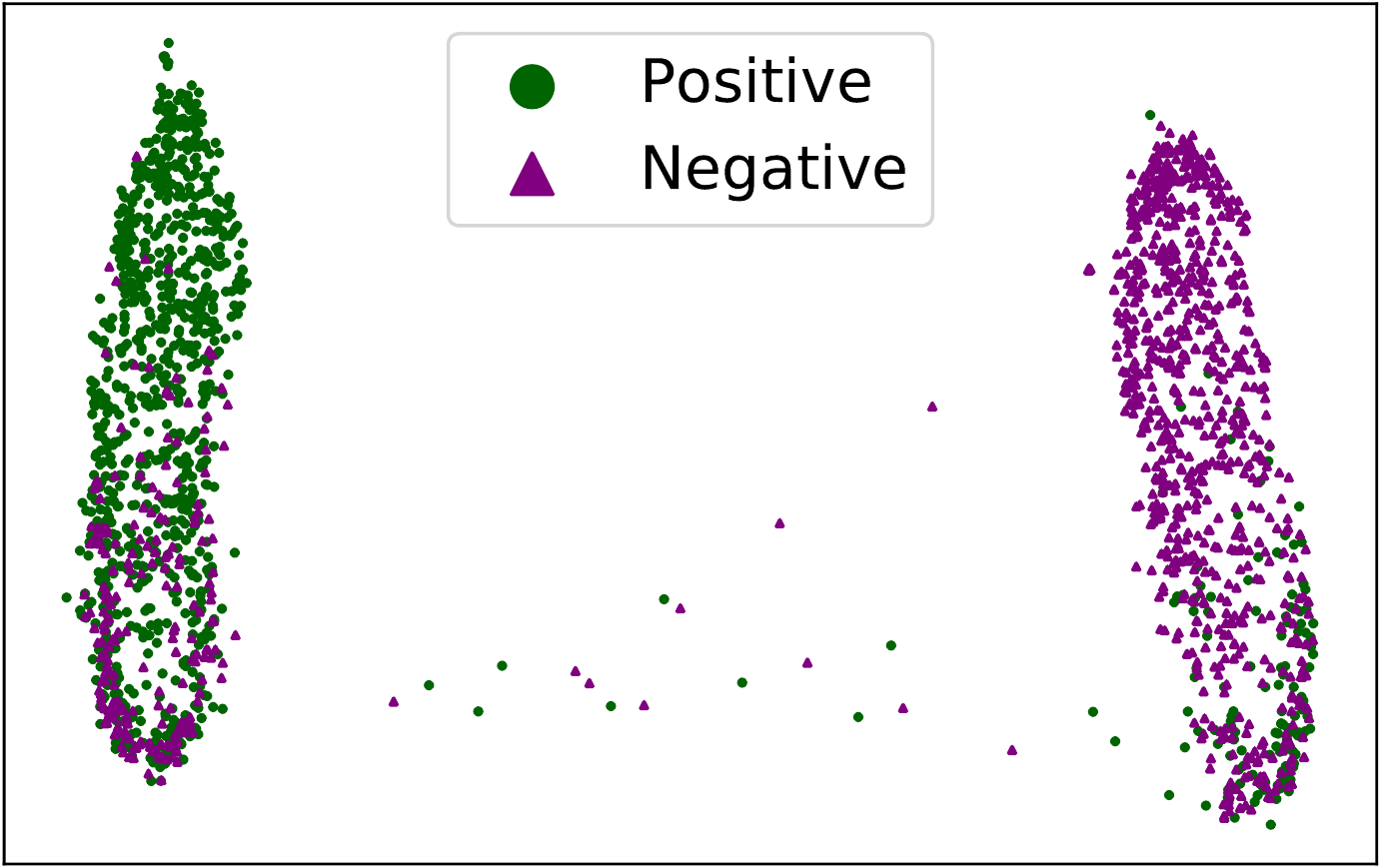} & \includegraphics[width=.24\linewidth,height=\height]{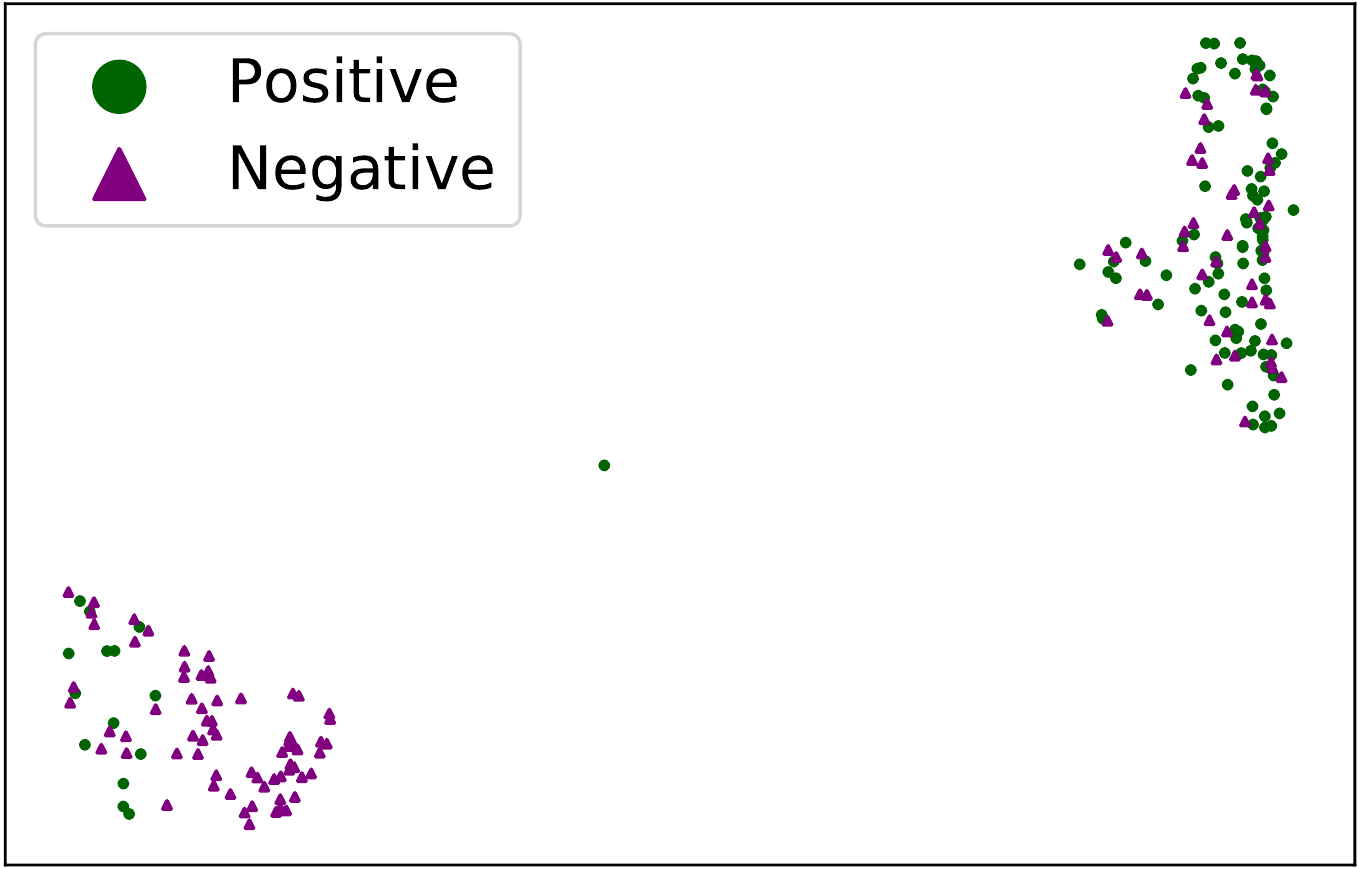} & \includegraphics[width=.24\linewidth,height=\height]{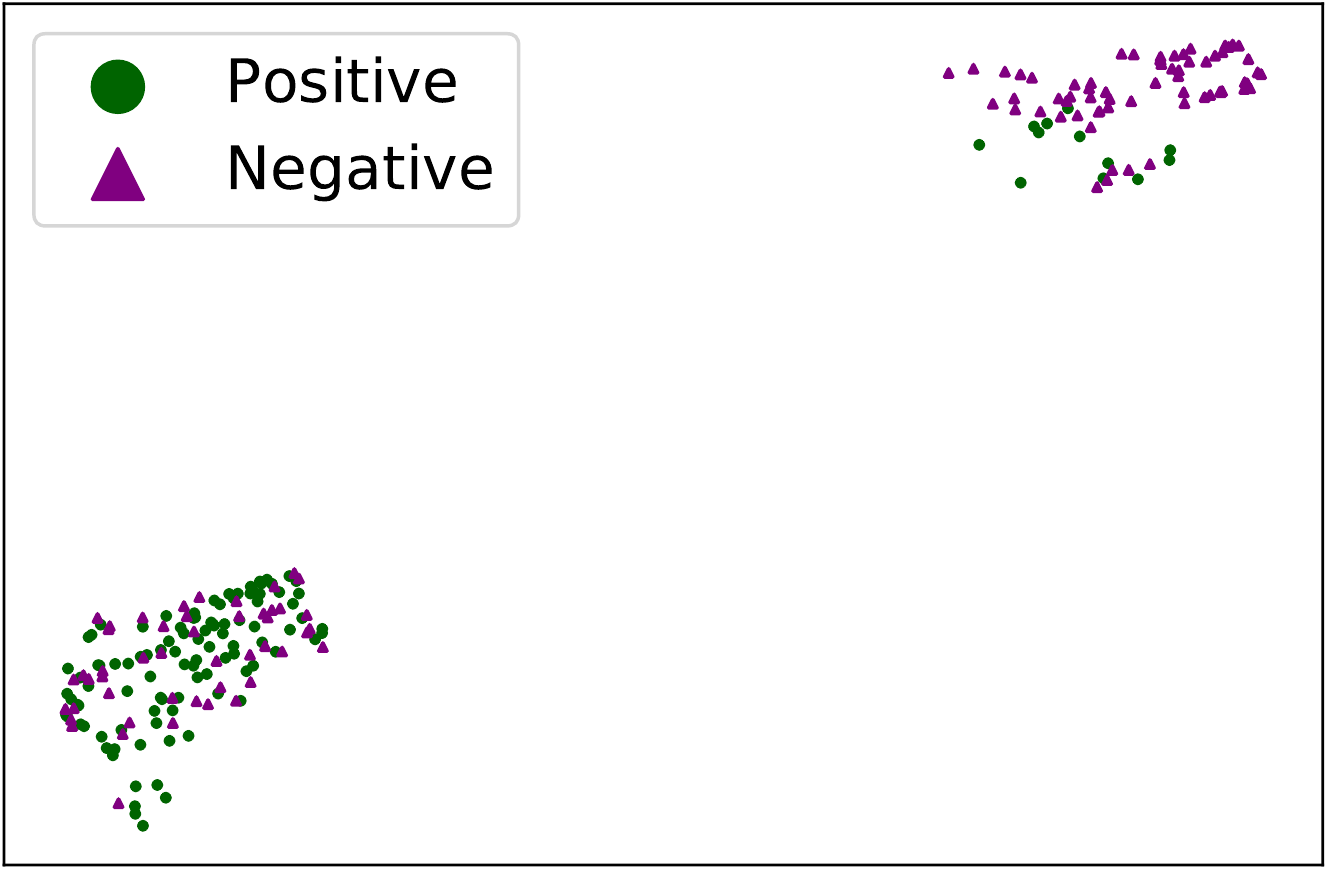} \\
    (a) 3\% -- FC-SVM & (d) 3\% -- XC-SVM  & (a) 3\% -- FC-SVM & (d) 3\% -- XC-SVM  \\
    \includegraphics[width=.24\linewidth,height=\height]{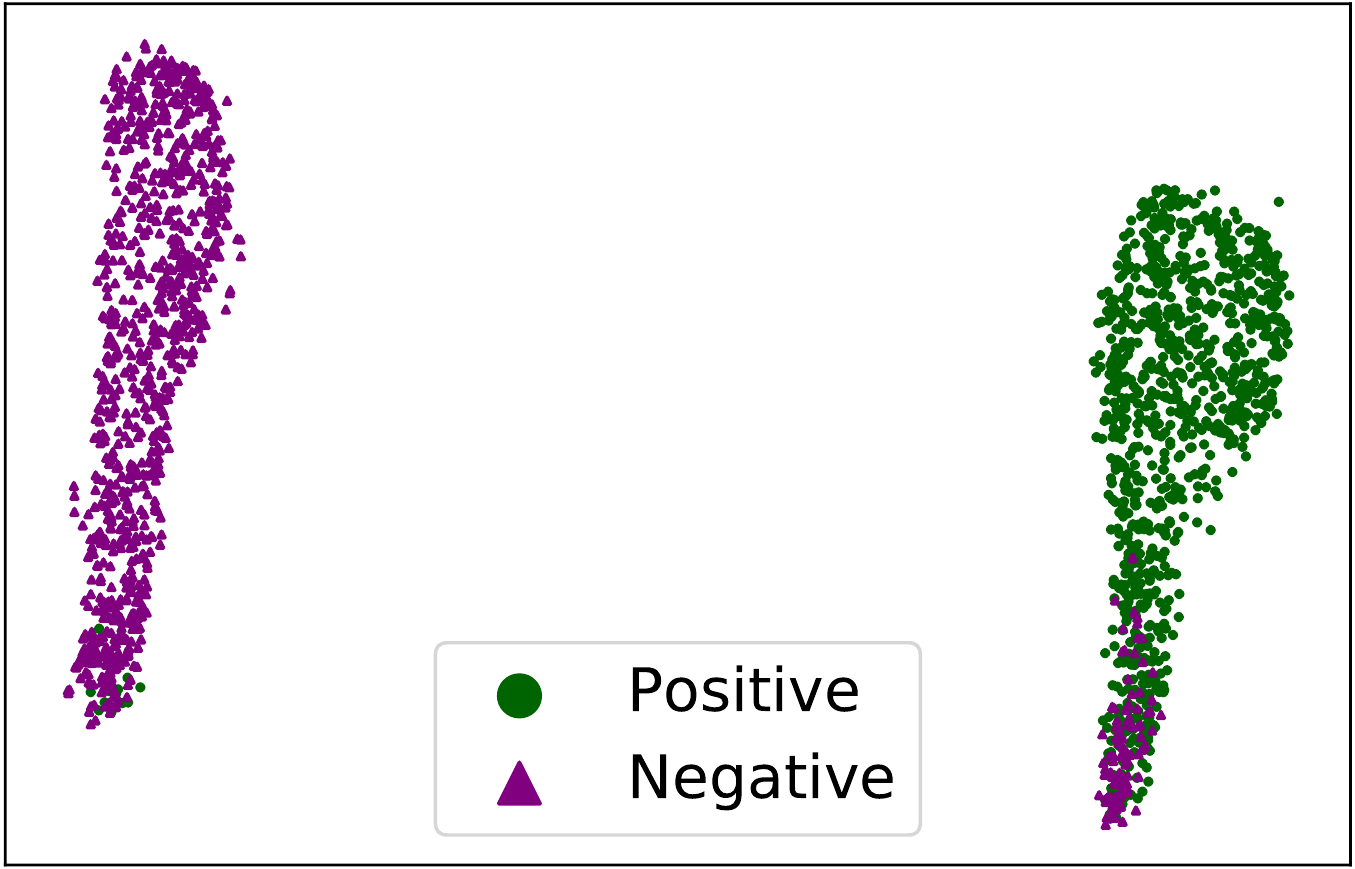} & \includegraphics[width=.24\linewidth,height=\height]{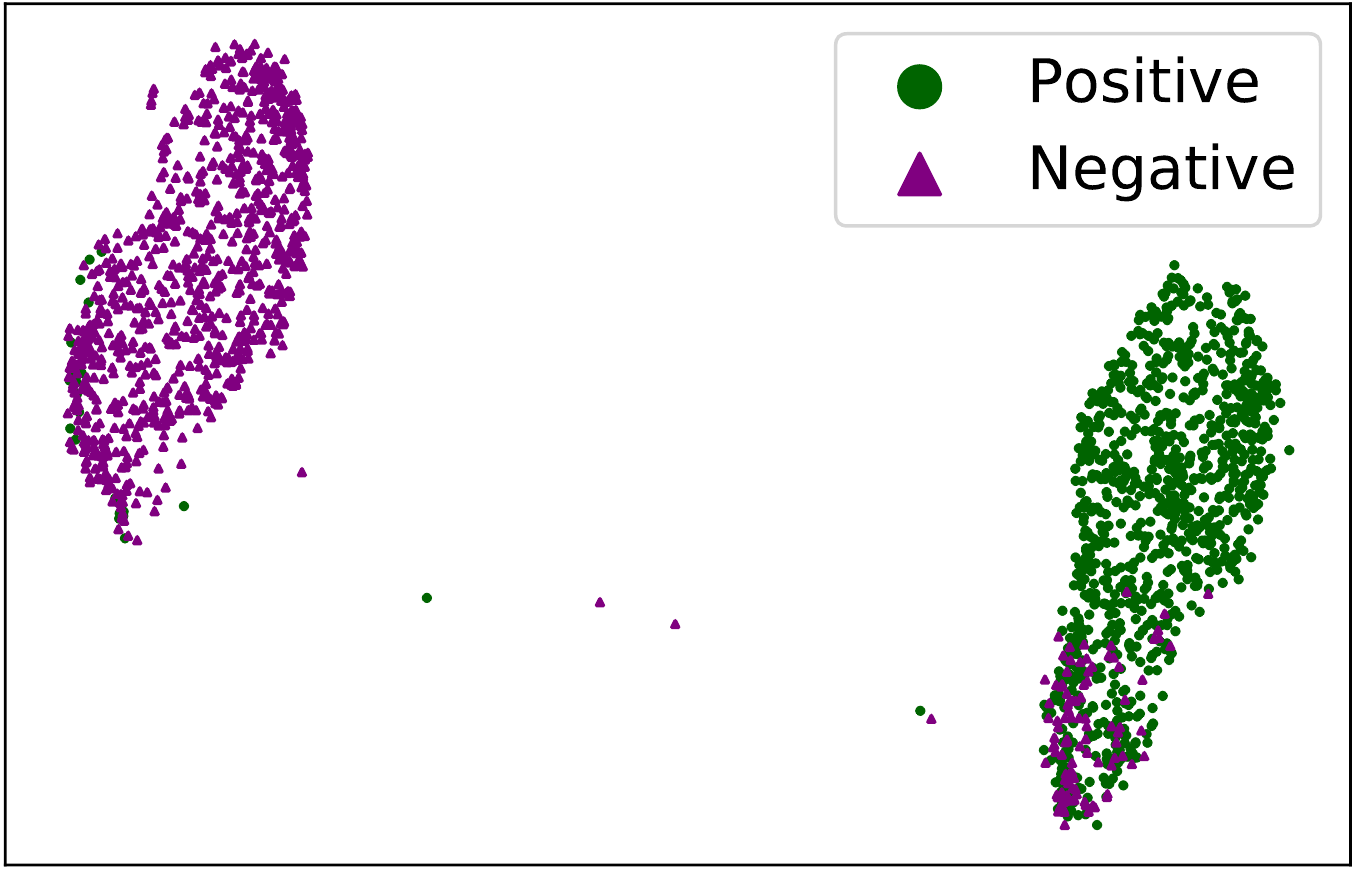} & \includegraphics[width=.24\linewidth,height=\height]{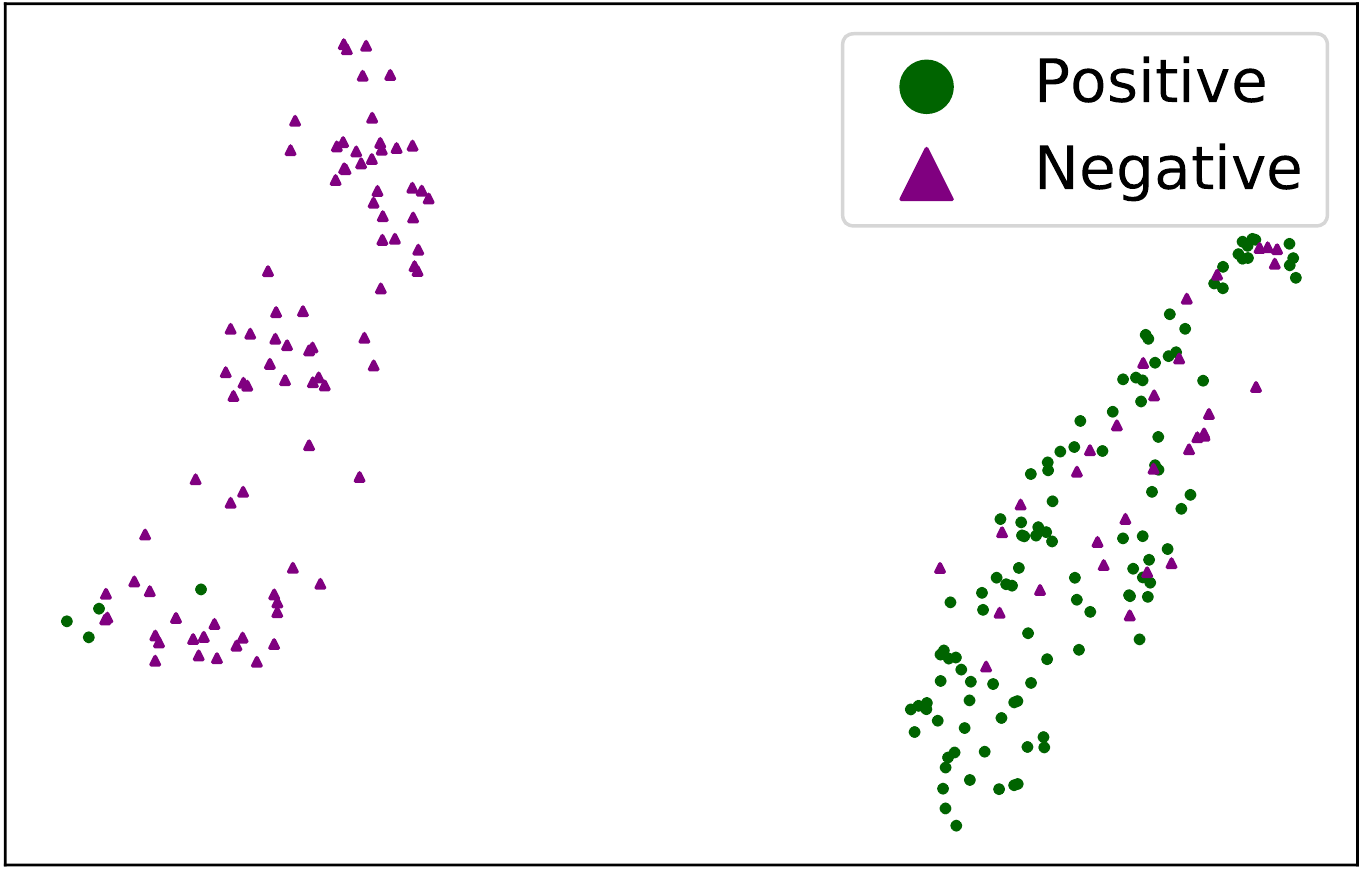} & \includegraphics[width=.24\linewidth,height=\height]{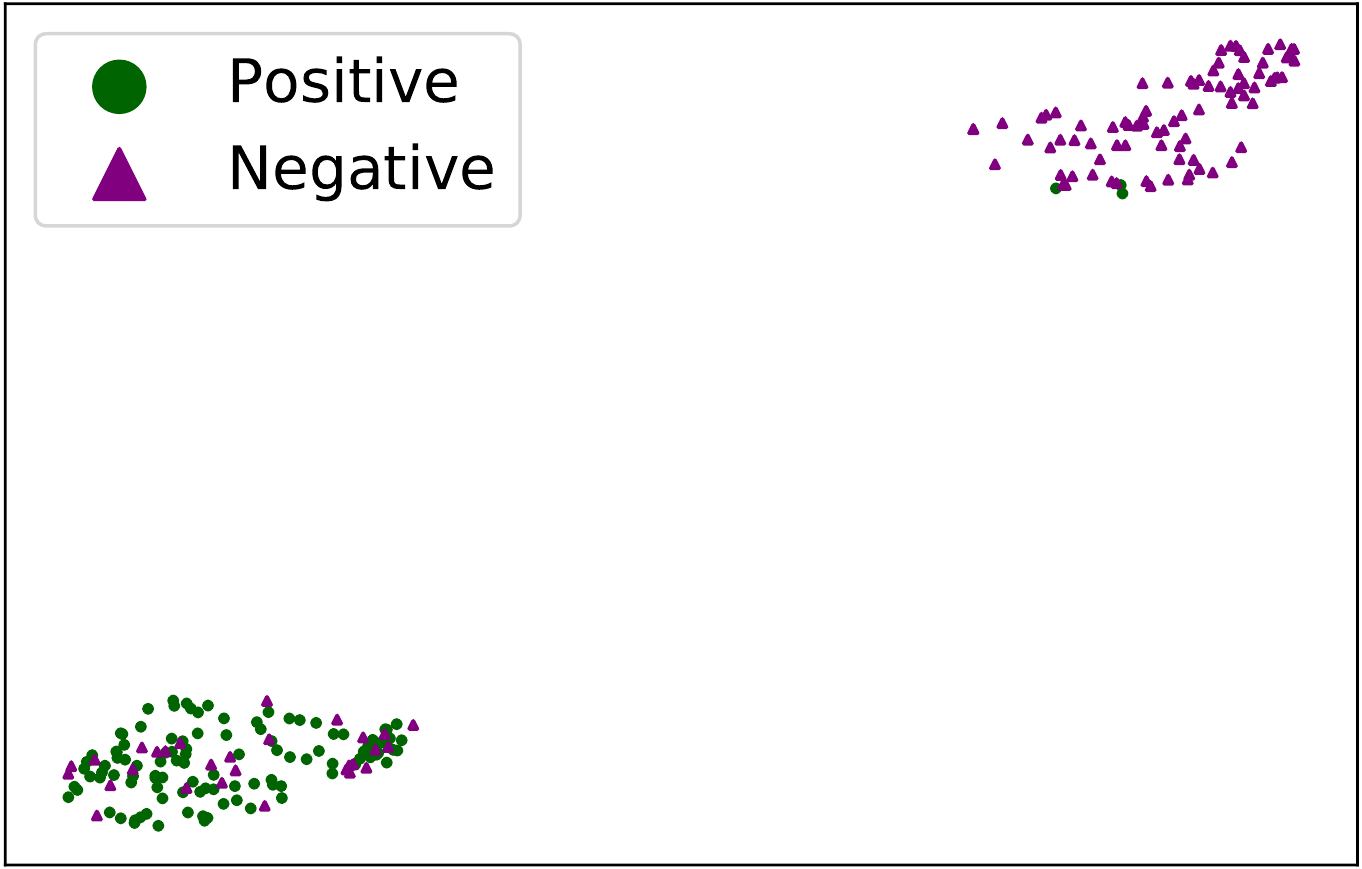} \\
    (b) 21\% -- FC-SVM & (e) 21\% -- XC-SVM & (b) 21\% -- FC-SVM & (e) 21\% -- XC-SVM \\
    \includegraphics[width=.24\linewidth,height=\height]{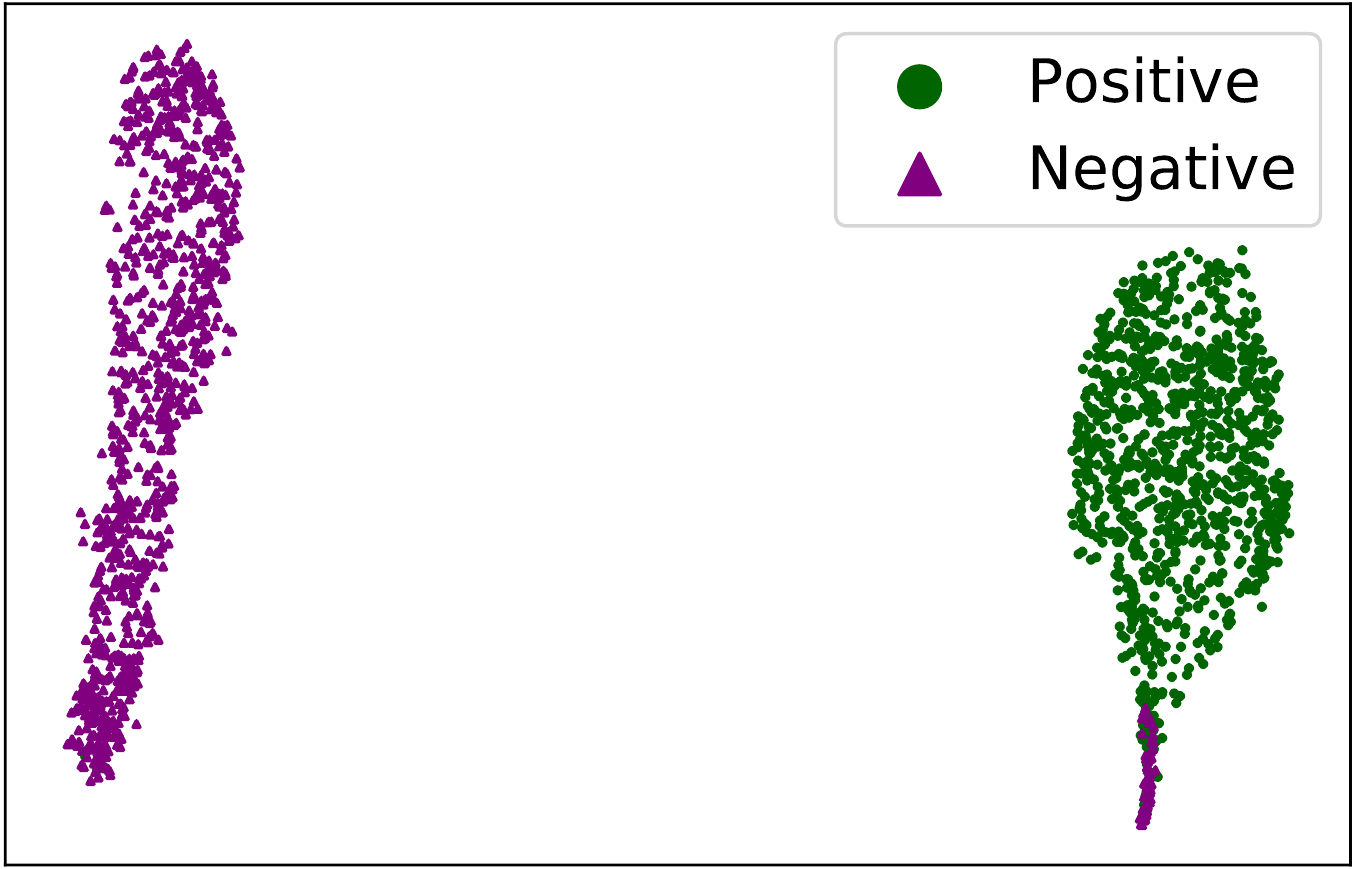} & \includegraphics[width=.24\linewidth,height=\height]{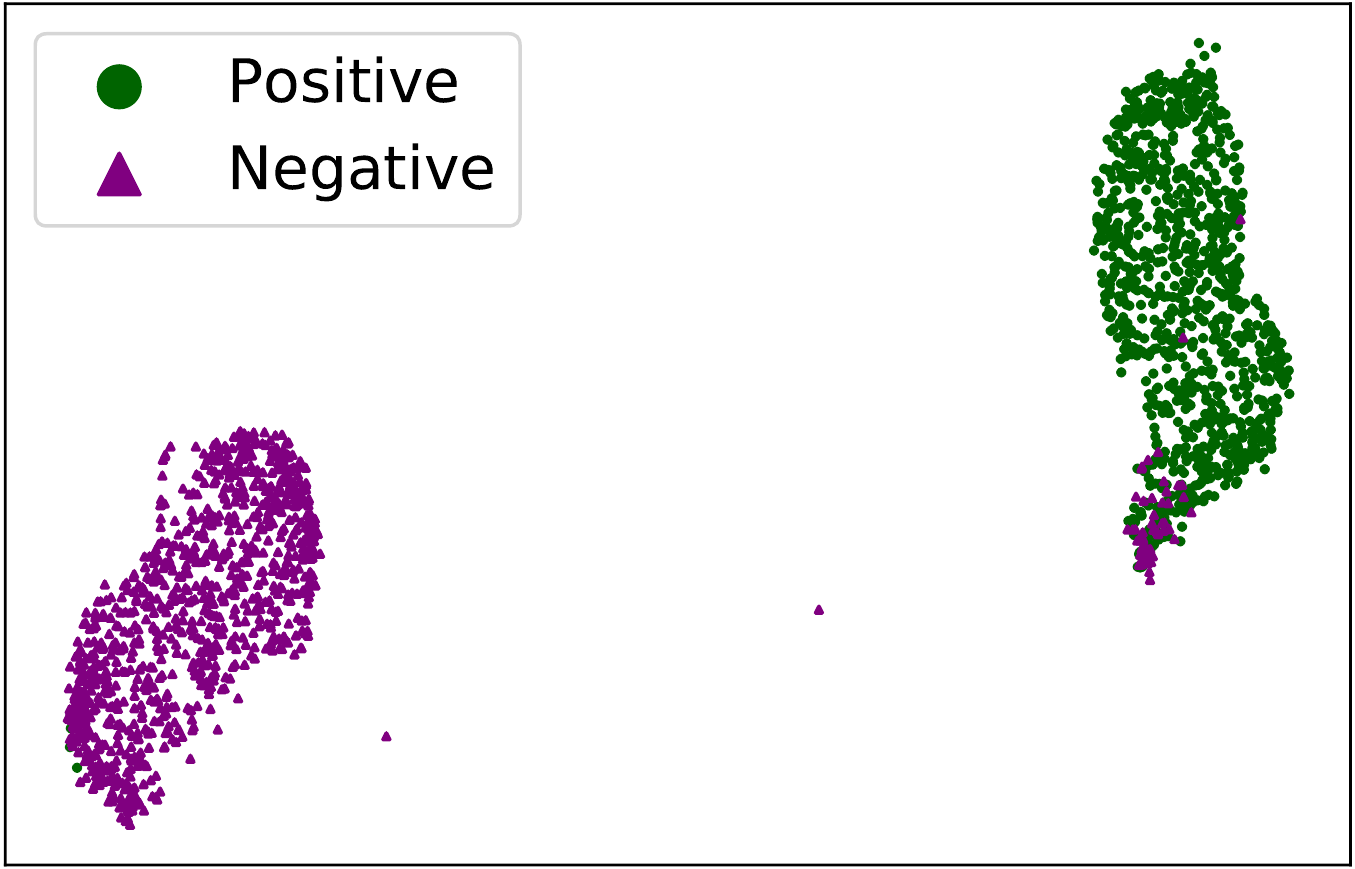} & \includegraphics[width=.24\linewidth,height=\height]{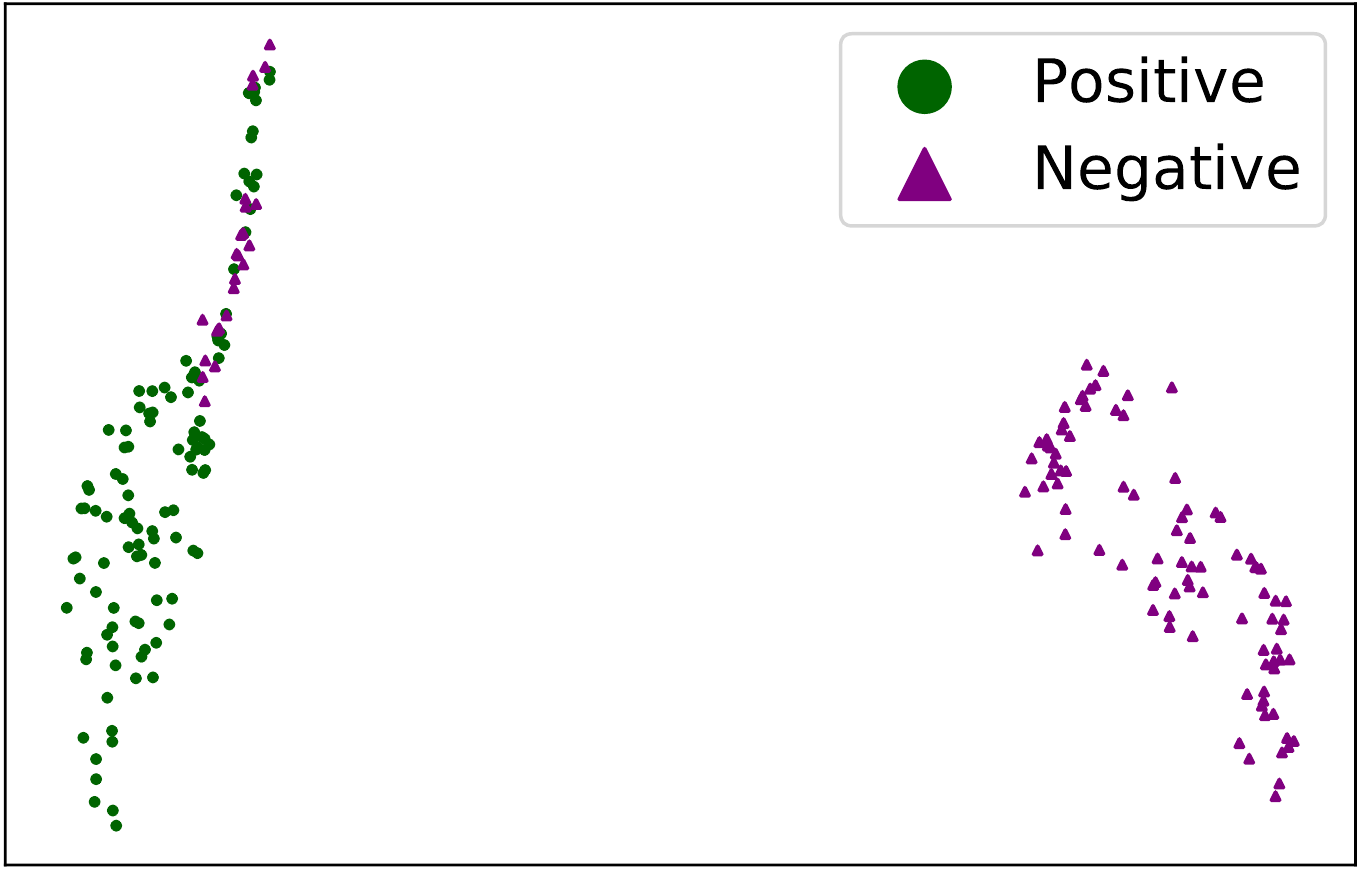} & \includegraphics[width=.24\linewidth,height=\height]{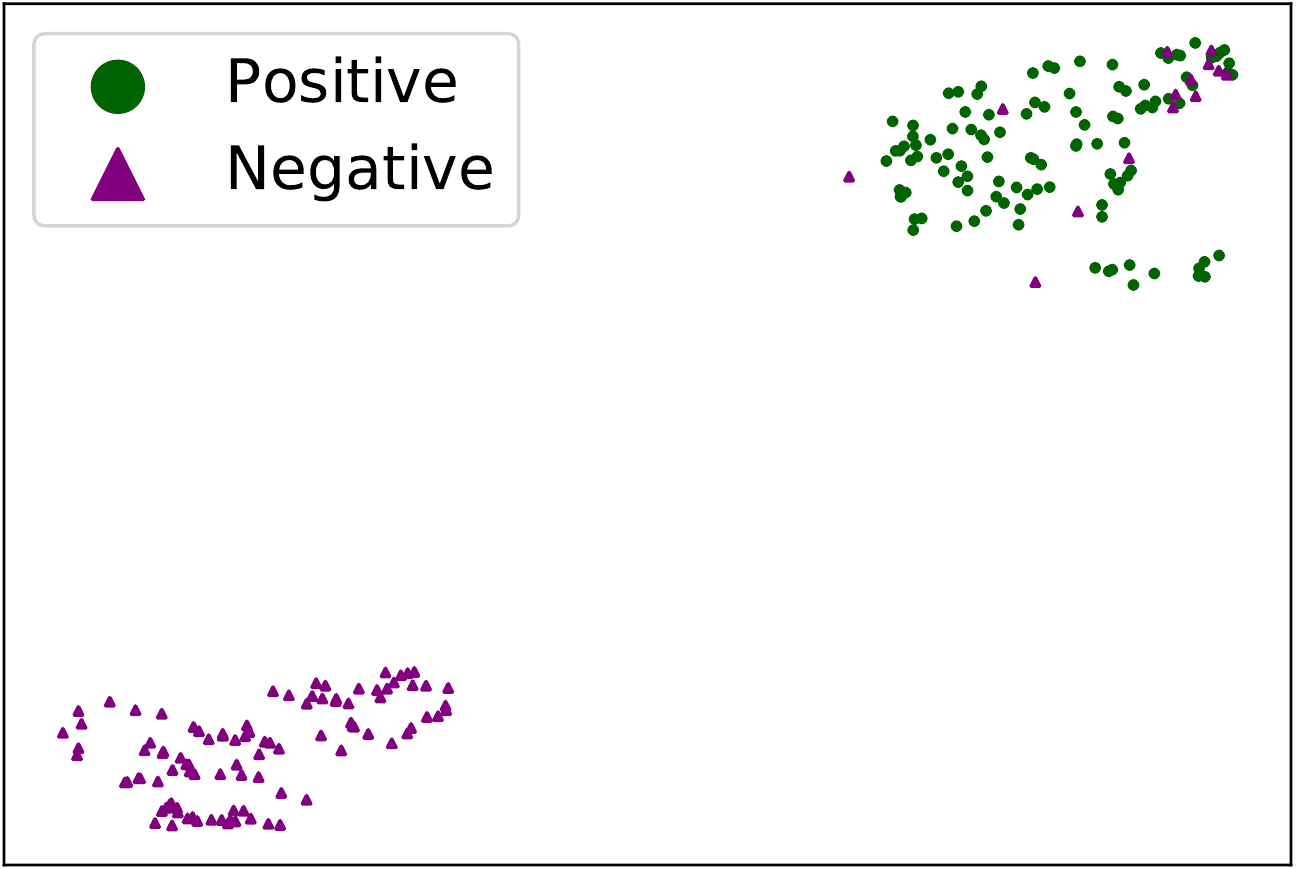} \\
   (c) 50\% -- FC-SVM & (f) 50\% -- XC-SVM & (c) 50\% -- FC-SVM & (f) 50\% -- XC-SVM \\
  \end{tabular}
  \caption{UMAP dimensionality reduction plots over SC-Image dataset (left) and Petrobras dataset (right), where ``positive" means that images were tampered, and ``negative" means that images were not attacked.}
  \label{f.umap}
\end{figure*}

All plots present two clusters as a consequence of the training of the network as a binary classifier. For images with a higher level of tampering, the clusters present smaller overlap among points from different classes. In such condition, the network presents a better discriminative power provided by the artifacts left by seam carving in the image. Such discriminative power is reduced with the decreasing of the tampering level, resulting in clusters with more misclassified samples. Besides, there are also points far away from clusters, which may reflect in a challenger situation for the classifiers.

\subsection{Petrobras Dataset}
\label{ss.petrobras_dataset}

Table~\ref{t.experiments_petrobras_dataset} presents the accuracy concerning the experiments over the Petrobras dataset for different levels of tampering, where the best results are in bold. Similarly to the previous dataset, images with lower levels of tampering exhibit smaller accuracy. Among the proposed approaches, XC-SVM obtained the best results in almost every distortion rates, except for $18\%$ and $30\%$. Regarding a change of rate as of 3\%, XC-SVM achieved the accuracy of $72.64\%$, followed by FC-SVM with $71.75\%$ of recognition rate. As the tampering level is intensified, all approaches also increase their accuracies. For images with $50\%$ of tampering, XC-SVM could reach the best accuracy of $95.45\%$. Such results corroborate with those from the other dataset.

The networks decision behavior can be easily visualized in Figure~\ref{f.umap} (right), depicting UMAP projections onto a bidimensional space of the output features from the XC-SVM and FC-SVM models. The projection output for $3\%$ of tampering presents two main clusters, where there is a moderate presence of mixing between classes. Such situation illustrates the confusion made by the network when images with a low level of tempering are evaluated. On the other hand, the projection of the 50\% carved images shows clearer cluster separation between classes. Although Petrobras and SC-Image datasets are different, their projections present strong similarities.

\begin{table}[!htb]
  \centering
  \caption{Accuracies obtained over Petrobras dataset.}
\label{t.experiments_petrobras_dataset}
  \begin{tabular}{p{2.4cm}rrr}
  \toprule
                    & \multicolumn{3}{c}{Accuracy (\%)}  \\ 
  Rate of Change    & SM-CNN         & FC-SVM          & XC-SVM          \\ \midrule
  3\%               & 67.86          & 71.75           & \textbf{72.64}  \\
  6\%               & 66.96          & 75.58           & \textbf{75.85}  \\
  9\%               & 76.79          & 77.90           & \textbf{77.99}  \\
  12\%              & 79.91          & 79.41           & \textbf{80.04}  \\
  15\%              & 81.25          & 82.27           & \textbf{83.78}  \\
  18\%              & \textbf{83.04} & 82.27           & 82.53           \\
  21\%              & 83.04          & 84.58           & \textbf{85.03}  \\
  30\%              & 84.82          & \textbf{90.02}  & 89.93           \\
  40\%              & 85.27          & 92.16           & \textbf{92.52}  \\
  50\%              & 85.71          & 94.57           & \textbf{95.45}  \\ \bottomrule
  \end{tabular}
\end{table}
 
\section{Conclusions}
\label{s.conclusions}

Seam carving is a resizing method widely used in image manipulation with minimal distortions aiming at preserving essential elements of the image's content.  Due to this characteristic, seam carving can be used to tamper with images by removing desired elements from the image.

In this paper, we proposed to detect image tampering using an end-to-end deep learning network. Experiments demonstrated that the proposed approach is suitable to be an auditing tool to identify seam-carved images since it can classify images with a low level of tampering with good accuracy, besides having an excellent performance for images with aggressive tampering. 

The projection of the output of the deep neural network features onto a bidimensional space indicates that the obtained clustered space might be suitable to build a hybrid-approach, i.e., we can attach another layer of classifiers right after the mapped space. The proposed approach opened new possibilities to explore new hybrid architectures. 

Concerning future works, we intend to evaluate the Optimum-Path Forest classifier~\cite{PapaIJIST:09,PapaPR:12,PapaPRL:17} in the context of seam carving detection since it is parameterless and fast for training. Besides,  we are considering employing deep autoencoders and subspace clustering~\cite{PanJi} for unsupervised tempering detection as well.


%
%
%
\bibliographystyle{splncs04}
\bibliography{references}
\end{document}